\documentclass[runningheads]{llncs}
\usepackage[T1]{fontenc}
\usepackage{graphicx}
\usepackage{amsmath}               
\usepackage{amssymb}  
\usepackage{graphicx}
\usepackage{subcaption}
\usepackage{algorithmic}
\usepackage{algorithm}

\DeclareMathOperator*{\argmin}{arg\,min}
\begin{document}
\title{Explore Beyond the Boundary Using Entropic Information}
%
%
\author{Bumgeun Park \and
Donghwan Lee}
\authorrunning{B. Park and D. Lee}
%
\institute{Korea Advanced Institute of Science and Technology, Daejeon 34051, Korea
\email{\{j4t123,donghwan\}@kaist.ac.kr}\\}
\maketitle              
\begin{abstract}
In reinforcement learning, exploration with sparse and delayed rewards presents a significant challenge due to the limited feedback available for guiding the learning process. Addressing this issue requires extensive exploration in the state space to discover valuable reward signals. In this paper, we propose Entropic Information for Exploration (ENTINEX), a novel method that enhances exploration by incentivizing agents to explore beyond the boundaries of the state distribution. ENTINEX achieves this by assigning intrinsic rewards to these boundaries, leveraging entropic information to identify them effectively. Through extensive experimentation, we demonstrate that ENTINEX consistently improves exploration performance in environments characterized by sparse and delayed rewards. Our experimental results show that ENTINEX outperforms existing exploration methods, highlighting its effectiveness in both sparse and delayed reward scenarios.

\end{abstract}
\section{Introduction}
\label{introduction}
The objective of reinforcement learning (RL) is to learn an optimal policy that maximizes cumulative rewards through interactions with an environment. Recent advances in RL have enabled substantial progress in various decision-making problems \cite{example_rl2}. However, in many real-world scenarios, sparse or delayed rewards provide insufficient feedback for effective policy learning, often leading to suboptimal performance \cite{sparse_reward2,delayed_reward1}.

Reward engineering can alleviate this problem by providing denser reward signals and has been applied to autonomous driving \cite{reward_autonomous}, healthcare \cite{reward_health}, traffic control \cite{reward_traffic}, and robot control \cite{reward_robot}. However, it requires substantial expertise and domain-specific knowledge, making it costly and difficult to generalize across tasks.

Intrinsic rewards have therefore been widely studied as a general approach to facilitating exploration. A straightforward approach is count-based exploration, which assigns a reward such as $\frac{1}{N(s)}$, where $N(s)$ denotes the visitation count of state $s$ \cite{count-based2}. This encourages the agent to visit less-explored states, but directly counting state visitations is impractical in continuous spaces. Alternative approaches estimate state novelty using prediction errors \cite{prediction-error2,prediction-error3}, state entropy \cite{re3}, or discretized feature counts \cite{discretized_feature}.

The boundary of the state-novelty distribution (SND), which separates explored and unexplored regions, provides an informative target for exploration. Boundary-based methods encourage agents to move beyond this boundary and thereby expand the explored region \cite{boundary_ref1,boundary_ref2,boundary_ref3,boundary_ref4}. Since state visitation generally decreases near the boundary, state-novelty and boundary-based approaches are closely related. For example, high-novelty states can be regarded as boundary states \cite{boundary_ref3}, while other methods explicitly estimate the boundary from the novelty difference between consecutive states \cite{boundary_ref1,boundary_ref2,boundary_ref4}.

In this paper, we propose \textbf{Ent}ropic \textbf{In}formation for \textbf{Ex}ploration (ENTINEX), an exploration method that assigns intrinsic rewards to states near the SND boundary. Unlike existing methods that rely on novelty differences between consecutive states, ENTINEX identifies the boundary using entropic information from the action probability distribution. Experimental results demonstrate that ENTINEX achieves promising improvements over existing exploration methods.

\begin{figure*}[t]
    \begin{center}
        \includegraphics[width=\textwidth]{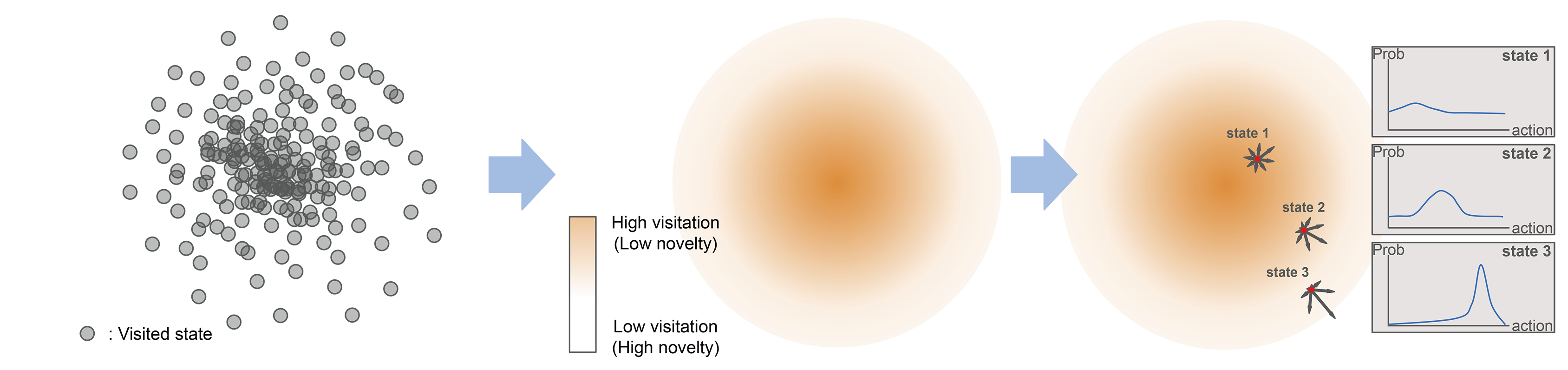}
    \end{center}
    \caption{Illustration of leveraging the entropic information. State visitation (left) is approximated by state-novelty distribution (center) using a novelty function. The novelty action probability distribution (right) is then computed, resulting in different distribution shapes depending on the state.}
    \label{procedure}
\end{figure*}

\section{Preliminaries}
\label{preliminaries}

\subsection{Problem Formulation}
\label{formulation}

We consider a Markov decision process (MDP) defined by the tuple $(\mathcal{S},\mathcal{A},P,r,\gamma)$, where $\mathcal{S}$ and $\mathcal{A}$ are the state and action spaces, $P(s'|s,a)$ is the transition probability, $r(s,a)$ is the reward function, and $\gamma\in[0,1)$ is the discount factor. At each time step $t$, the agent selects an action $a_t\sim\pi(\cdot|s_t)$ and transitions to $s_{t+1}\sim P(\cdot|s_t,a_t)$. The objective is to learn a policy that maximizes the expected cumulative discounted reward:
\begin{equation}
J(\pi)=\sum_{t=0}^{\infty}\mathbb{E}\bigg[\gamma^{t}r(s_t,a_t)\bigg|\pi\bigg].
\end{equation}
We consider continuous state and action spaces.

\subsection{State Novelty}
\label{state_novelty}
State novelty can be used to construct intrinsic rewards for hard exploration tasks.

\begin{definition}[State Novelty]
Let $\phi:\mathcal{S}\rightarrow\mathbb{R}$ denote the novelty of a state $s\in\mathcal{S}$.
\end{definition}

In discrete spaces, $\phi(s)$ is typically defined using inverse state-visitation counts \cite{count-based2}. In continuous spaces, it can be estimated using prediction errors \cite{prediction-error2,prediction-error3} or state entropy \cite{re3}.

\begin{definition}[Expected Novelty]
Let $\phi':\mathcal{S}\times\mathcal{A}\rightarrow\mathbb{R}$ denote the expected novelty of the next state:
\begin{equation}
\phi'(s,a)=\mathbb{E}_{s'\sim P(\cdot|s,a)}[\phi(s')].
\end{equation}
\end{definition}

\begin{definition}[Novelty Action Probability Distribution]
Let $\psi(\cdot|s)$ denote the novelty action probability distribution defined by
\begin{equation}
\psi(a|s)=
\frac{\exp\big(\phi'(s,a)\big)}
{\int_{\mathcal{A}}\exp\big(\phi'(s,a')\big)da'}.
\label{napd}
\end{equation}
\end{definition}

We denote an action sampled from $\psi$ by $a_\psi$ to distinguish it from an action sampled from the policy $\pi$.

\section{Related works}
\label{related_work}

\subsection{Exploration in RL}
\label{exploration_in_rl}
Exploration is a fundamental problem in RL because sufficient visitation of state-action pairs is required for reliable value estimation and convergence \cite{q-function}. Simple exploration strategies include $\epsilon$-greedy action selection \cite{dqn,double_dqn}, action noise \cite{ddpg}, and entropy regularization, which promotes diverse behaviors \cite{sac}.

More advanced methods guide agents toward unexplored regions by assigning intrinsic rewards to rarely visited states or state-action pairs. In discrete spaces, these rewards can be derived directly from visitation counts \cite{count-based2}. In continuous spaces, where direct counting is impractical, state novelty is commonly estimated using prediction errors \cite{prediction-error2,prediction-error3} or state entropy \cite{re3}. For example, RE3 \cite{re3} estimates state entropy without training an additional model and achieves competitive exploration performance.

\subsection{Entropy for Measuring Uncertainty}
Entropy is widely used to quantify uncertainty and information in probability distributions. In active learning, high-entropy samples are prioritized for labeling because they indicate predictive uncertainty \cite{active_learning1}. Entropy has also been used to characterize semantic uncertainty in natural language processing \cite{nlp_entropy2} and uncertainty in robot states and maps \cite{robot_entropy1}.

In RL, entropy is primarily used to represent policy uncertainty. Maximum-entropy methods augment the RL objective with policy entropy to encourage diverse actions and robust behaviors \cite{sac}. Entropy can also quantify state uncertainty or novelty and has therefore been incorporated into exploration methods \cite{re3}.

\section{Method}
\label{method}
\label{entinex}
\begin{figure}[t]
    \begin{center}
        \includegraphics[width=0.4\textwidth]{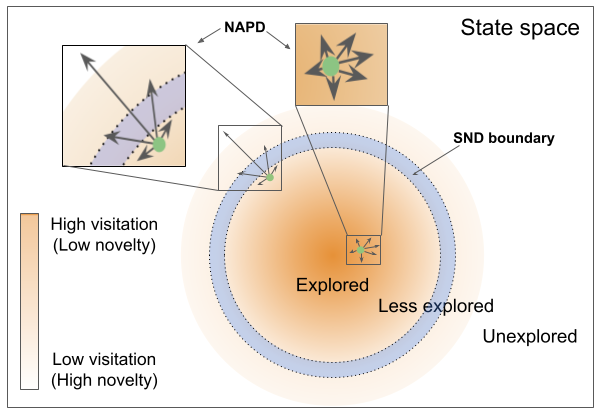}
    \end{center}
\caption{Illustration of SND boundary and NAPD. The SND boundary represents the area between explored and unexplored states. The amount of exploration is evaluated by the novelty function, expressed by the color. The NAPD represents the probability distribution over actions. Given the state expressed by the green dot, each arrow corresponds to a possible action and its length represents the probability of sampling that action.}
\label{overall_concept}
\end{figure}

State-distribution boundaries separate explored and unexplored regions and therefore provide useful targets for discovering novel states. Previous methods define boundaries using high-novelty states \cite{boundary_ref3} or large representation or novelty differences between consecutive states \cite{boundary_ref1,boundary_ref2,prediction-error2}. Based on this perspective, we identify boundary states using the entropy of an action distribution induced by state novelty.

\subsection{Entropic Information for Exploration}
We propose \textbf{Ent}ropic \textbf{In}formation for \textbf{Ex}ploration (ENTINEX), an intrinsic-reward method for environments with sparse or delayed rewards. ENTINEX uses the novelty action probability distribution (NAPD), a Boltzmann distribution induced by the expected novelty of the next state. The NAPD assigns higher probabilities to actions leading to more novel states.

Near the state-novelty distribution (SND) boundary, some actions may lead beyond the explored region and thus have substantially higher expected novelty, producing a biased, low-entropy NAPD. Far from the boundary, actions tend to have similar expected novelty, resulting in a more uniform, high-entropy distribution. ENTINEX therefore uses the negative entropy of the NAPD to identify and reward states near the SND boundary. Figures~\ref{overall_concept} and \ref{procedure} illustrate this concept and its implementation.

\subsubsection{State-Novelty Distribution}
Direct state-visitation counting is impractical in high-dimensional or continuous spaces. We instead approximate the state distribution using the state-novelty distribution (SND), induced by a state novelty function, as illustrated in Fig.~\ref{procedure} (center).

\subsubsection{Novelty Action Probability Distribution}
The NAPD is the Boltzmann distribution defined in \eqref{napd} using the expected novelty $\phi'(s,a)$. It becomes biased when a particular action leads to substantially higher novelty and more uniform when actions have similar expected novelty. Its entropy can therefore indicate whether a state lies near the SND boundary.

Directly following actions sampled from the NAPD would encourage only one-step novelty. Instead, ENTINEX uses its entropic information as an intrinsic reward, allowing the policy to maximize novelty over long-term trajectories. The following theorem relates the NAPD entropy to the consecutive-state novelty difference used by previous boundary-based methods.
\begin{theorem}
\label{theorem_boundary_approximation}
Let $s\in\mathcal{S}$ be a state near the SND boundary. Then,
\begin{equation}
\begin{aligned}
-\mathcal{H}\big(\psi(\cdot|s)\big)
\simeq
\mathbb{E}_{a\sim\psi(\cdot|s)}
\bigg[
\phi'(s,a)-\phi(s)
-\max_{a'}\big(\phi'(s,a')-\phi(s)\big)
\bigg].
\end{aligned}
\label{theorem1}
\end{equation}
\end{theorem}
Here, $\mathcal{H}$ denotes entropy. Equation~\eqref{theorem1} relates the negative NAPD entropy to the consecutive-state novelty difference, normalized by its maximum value.

Previous methods estimate boundaries using novelty differences for actions sampled from a behavior policy or replay buffer \cite{boundary_ref1,boundary_ref2,boundary_ref4,prediction-error2}. In contrast, \eqref{theorem1} uses actions sampled from $\psi(\cdot|s)$. The discrepancy caused by these different action distributions is bounded as follows.
\begin{theorem}
\label{theorem_boundary_discrepancy}
For action distributions $\psi(\cdot|s)$ and $p(\cdot|s)$,
\begin{equation}
\begin{aligned}
\left|\mathbb{E}_{a\sim\psi(\cdot|s)}[\phi'(s,a)-\phi(s)]-\mathbb{E}_{a\sim p(\cdot|s)}[\phi'(s,a)-\phi(s)]
\right|\quad\quad\quad\quad\quad\quad\quad
\\\quad\quad\quad\quad\quad\quad\quad\leq
2D_{TV}\big(\psi(\cdot|s),p(\cdot|s)\big)
\big|\phi(s)+\epsilon_{\phi}(s)\big|.
\end{aligned}
\label{theorem2}
\end{equation}
\end{theorem}
Here,
\begin{equation}
D_{TV}(p,q)
=
\frac{1}{2}\int_{\mathcal{A}}|p(a)-q(a)|,da
\end{equation}
is the total variation distance, and
$\epsilon_{\phi}(s)=\max_a|\phi'(s,a)-\phi(s)|$.


\subsection{Practical Implementation of ENTINEX}
We implement ENTINEX with an off-policy RL algorithm for sample efficiency. Directly computing
\begin{equation}
-\mathcal{H}(\psi(\cdot|s))
=
\mathbb{E}_{a_\psi\sim\psi(\cdot|s)}
[\log\psi(a_\psi|s)]
\end{equation}
is difficult in high-dimensional continuous spaces. We therefore use the sample-based intrinsic reward
\begin{equation}
r^{\mathrm{int}}(s)
=
\log\psi(a_\psi|s),
\qquad
a_\psi\sim\psi(\cdot|s).
\end{equation}
The total reward is
\begin{equation}
r^{\mathrm{total}}_t
=
r(s_t,a_t)+\alpha r^{\mathrm{int}}(s_t),
\end{equation}
where $\alpha$ controls the intrinsic-reward weight. The complete procedure is provided in Algorithm~\ref{proposed_method}.

\subsubsection{Novelty Action Probability Distribution (NAPD)}
The normalizing integral of the Boltzmann NAPD is intractable in continuous action spaces. We therefore project it onto a family $\Pi$ of parameterized Gaussian distributions by minimizing
\begin{equation}
\psi(\cdot|s)
=
\argmin_{\psi'\in\Pi}
D_{KL}
\left(
\psi'(\cdot|s)
\middle\Vert
\frac{\exp\big(\phi'(s,\cdot)\big)}
{\int_{\mathcal{A}}\exp\big(\phi'(s,a)\big)da}
\right).
\end{equation}
Because the normalizing term is independent of the parameters of $\psi$, the resulting loss is
\begin{equation}
\mathcal{L}_{\psi}
=
\mathbb{E}_{s\sim\mathcal{D}}
\left[
\mathbb{E}_{a_\psi\sim\psi(\cdot|s)}
\left[
\log\psi(a_\psi|s)-\phi'(s,a_\psi)
\right]
\right],
\label{action_distribution_loss}
\end{equation}
where $\mathcal{D}$ denotes the replay buffer.

\subsubsection{Dynamics Model Learning}
Updating the NAPD requires $\phi'(s,a_\psi)$ for actions sampled from $\psi$, which may not have been executed in the environment. We therefore use a learned dynamics model $\hat{P}$ to predict the corresponding next state without additional environment interactions.

Following \cite{dynamics_model}, we use a latent dynamics model consisting of an encoder and decoder. The dynamics model is trained jointly with the policy rather than pre-trained, following the concurrent training strategy used in \cite{gobi}. Its loss is
\begin{equation}
\mathcal{L}_{\hat{P}}
=
\left|
d_{\omega}\big(e_{\omega}(s_t),a_t\big)
-
e_{\omega^{-}}(s_{t+1})
\right|_2^2,
\label{dynamics_model_loss}
\end{equation}
where $e_\omega$ and $d_\omega$ are the encoder and decoder, respectively, and $\omega^{-}$ is a slow-moving average of $\omega$.

\subsubsection{State Novelty Function}
Because direct visitation counting does not scale to continuous or high-dimensional state spaces, novelty can instead be estimated using density models \cite{density_model} or random encoders \cite{re3,prediction-error3}. We use the random encoder of RE3 \cite{re3} as the state novelty function $\phi(s)$. It estimates novelty through state entropy without requiring model updates during training. ENTINEX can also incorporate a learnable novelty estimator by adding its update step after minibatch sampling in Algorithm~\ref{proposed_method}.

\begin{algorithm}[tb!]
\caption{ENTINEX}
\label{proposed_method}
\begin{algorithmic}[1] 
\STATE Initialize parameters of state novelty function $(\phi)$, novelty action probability distribution $(\psi)$, dynamics model $(\hat{P})$
\STATE Initialize replay memory $\mathcal{M}\gets\emptyset$
\FOR{each step $t$}
\STATE Collect a transition $(s_t,a_t,r_t,s_{t+1})$ from environment using policy $\pi$
\STATE $\mathcal{M}\gets\mathcal{M}\cup\{(s_t,a_t,r_t,s_{t+1})\}$
\STATE Sample random minibatch $\{(s_i,a_i,a'_i,r_i)\}_{i=1}^{M}\sim\mathcal{M}$
\STATE Minimize \eqref{dynamics_model_loss}.
\STATE Minimize \eqref{action_distribution_loss}.
\FOR{$i=1$ \textbf{to} $M$}
\STATE Sample action $a_\psi\sim\psi(\cdot|s_i)$
\STATE Compute $r_i^{int}\gets\log\psi(a_\psi|s_i)$
\STATE Let $r_i^{total}\gets r_{i}+r_{i}^{int}$
\ENDFOR
\STATE Update policy with modified random minibatch $\{(s_i,a_r,s'_i,r_i^{total})\}_{i=1}^M$
\ENDFOR
\end{algorithmic}
\end{algorithm}

\begin{figure*}[hbt!]
    \centering
    \begin{subfigure}[b]{\textwidth}
        \centering
        \includegraphics[width=0.7\textwidth]{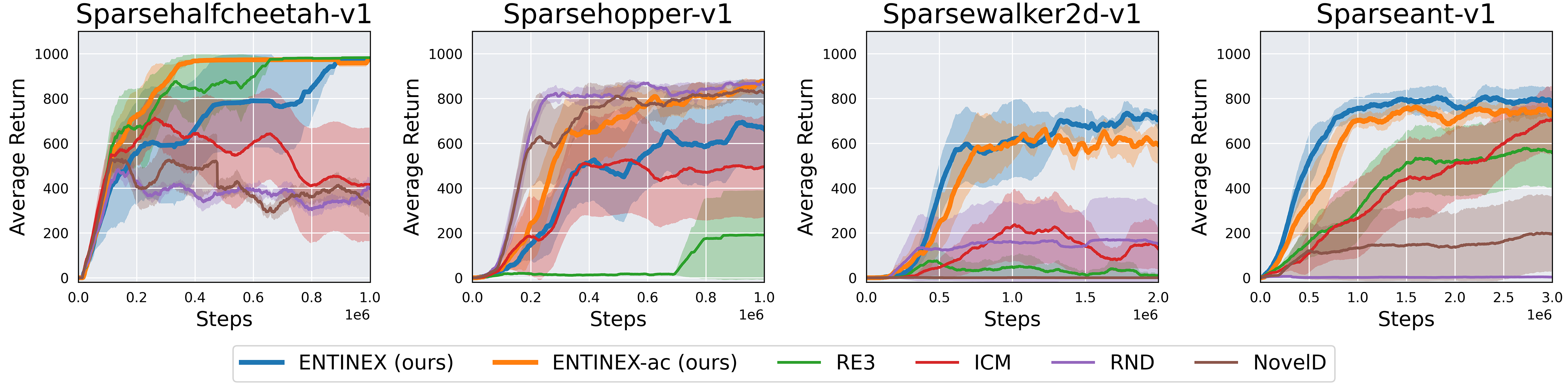}\label{sparse_result}
        \caption{Exploration performance evaluation on sparse reward environments}
        \label{sparse_result}
    \end{subfigure}
    \hfill
    \begin{subfigure}[b]{\textwidth}
        \centering
        \includegraphics[width=0.7\textwidth]{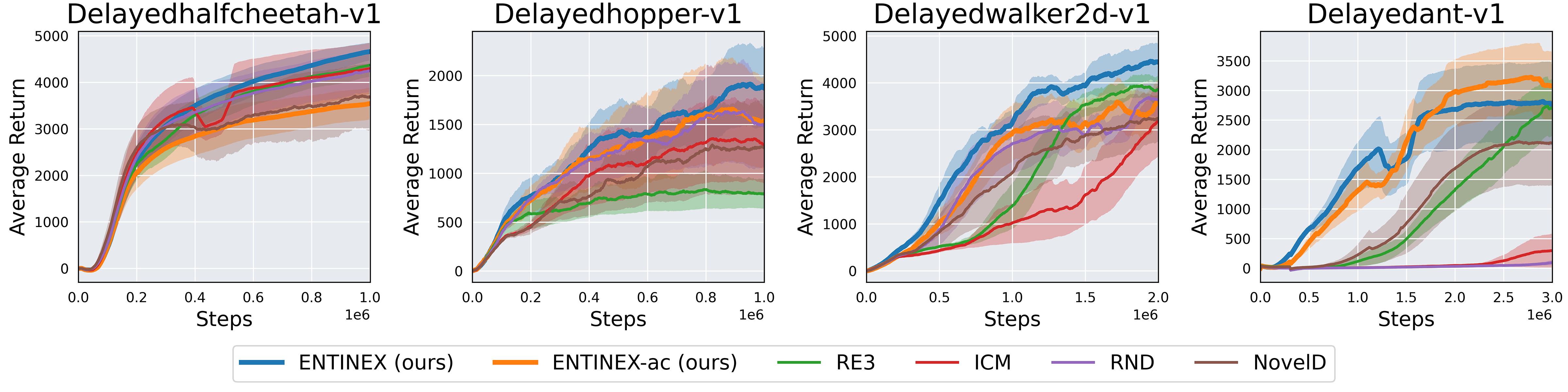}\label{delayed_result}
        \caption{Exploration performance evaluation on delayed reward environments}
        \label{delayed_result}
    \end{subfigure}
    \caption{Performance on MuJoCo environments with sparse and delayed reward. Each curve represents the average return across 5 different random seeds. The width of the shaded area represents 0.5 standard deviation $(0.5\sigma)$.}
    \label{result}
\end{figure*}

\section{Experiments}
\label{experiments}
Our experiments address the following questions:
\begin{itemize}
\item Does ENTINEX outperform existing exploration methods under sparse and delayed rewards?
\item Is its improvement over NovelD \cite{boundary_ref4} attributable to the entropic information-based approach rather than the novelty measure?
\item Does ENTINEX remain effective with different state novelty functions?
\item Can unsupervised pre-training with ENTINEX improve the sample efficiency of off-policy RL?
\end{itemize}

\begin{figure*}[hbt!]
\centering
\begin{subfigure}[b]{\textwidth}
    \centering
    \includegraphics[width=0.7\textwidth]{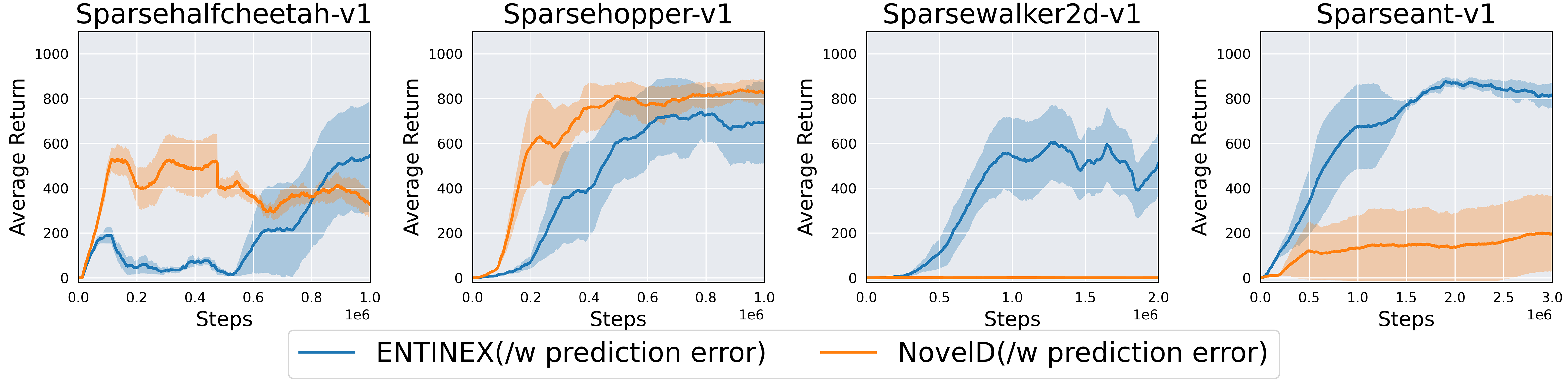}\label{novelty_compared_sparse_result}
    \caption{Comparison with another boundary-based method using the same novelty measure on sparse reward environments}
    \label{novelty_compared_sparse_result}
\end{subfigure}
\hfill
\begin{subfigure}[b]{\textwidth}
    \centering
    \includegraphics[width=0.7\textwidth]{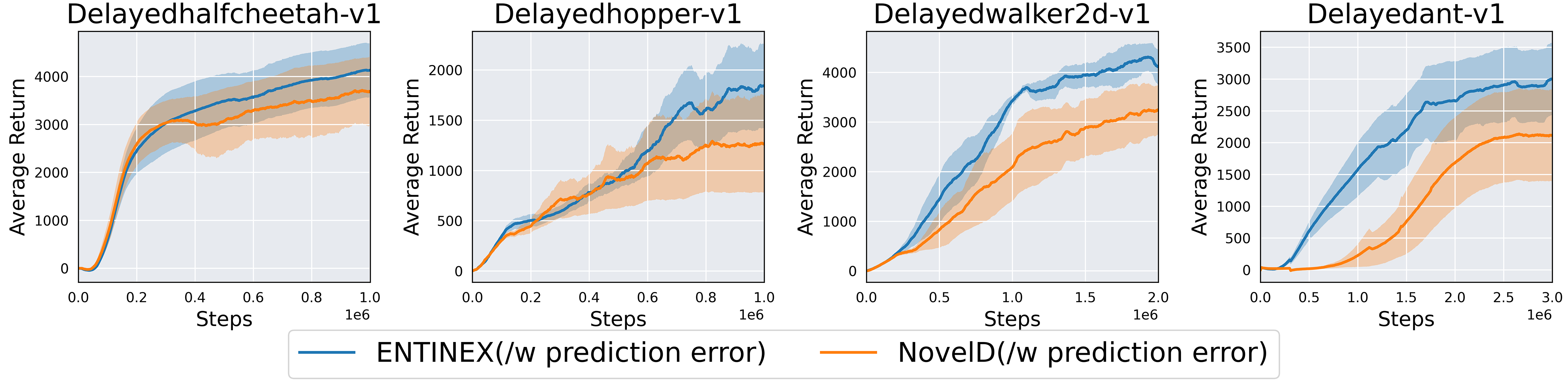}\label{novelty_compared_delayed_result}
    \caption{Comparison with another boundary-based method using the same novelty measure on delayed reward environments}
    \label{novelty_compared_delayed_result}
\end{subfigure}
\caption{Performance comparison with another boundary-based method, NovelD, to clarify that the superior performance of ENTINEX comes from not only novelty measure but also its approach to identifying the boundary}
\label{novelty_compared_result}
\end{figure*}

\subsection{Experimental Setup}
We compare ENTINEX with representative intrinsic-reward methods based on prediction error, including ICM \cite{prediction-error2} and RND \cite{prediction-error3}; state entropy, including RE3 \cite{re3}; and state-distribution boundaries, including NovelD \cite{boundary_ref4}. We use SAC as the underlying algorithm for all methods to isolate the effect of their intrinsic rewards, although their original implementations may use different algorithms. We also evaluate ENTINEX-ac, which uses a vanilla actor-critic algorithm.

We evaluate exploration performance on continuous-control MuJoCo environments \cite{mujoco} from OpenAI Gym \cite{openai}, modified to have sparse or delayed rewards. In the sparse-reward setting, the agent receives a reward of $+1$ only when a predefined condition is satisfied, following \cite{mujoco_sparse1,mujoco_sparse2,sparse_env}. In the delayed-reward setting, the original rewards are accumulated and delivered every $T$ time steps.

Directly visualizing exploration boundaries in high-dimensional state spaces is challenging. Standard dimensionality-reduction methods primarily capture sample similarity or density and may not reliably represent spatial relationships and visitation frequencies. We therefore use sparse-reward tasks in which rewards are obtained only by reaching positions far from the initial state. Performance on these tasks serves as an empirical proxy for boundary exploration.

\subsection{Results}
\subsubsection{Performance under Sparse and Delayed Rewards}
We compare ENTINEX with the baseline methods under sparse and delayed rewards. As shown in Fig.~\ref{result}, ENTINEX performs consistently across all sparse-reward environments. In SparseWalker2d, ENTINEX and ENTINEX-ac achieve final average returns of 705.7 and 596.2, respectively, while the baselines fail to learn effectively. ENTINEX also outperforms the baselines across all delayed-reward environments. These results demonstrate its effectiveness under both sparse and delayed feedback.

\subsubsection{Comparison with NovelD under a Matched Novelty Measure}
Both ENTINEX and NovelD identify exploration boundaries using state novelty. However, ENTINEX uses entropic information with the state-entropy estimator of RE3 \cite{re3}, whereas NovelD uses consecutive-state novelty differences with the prediction-error estimator of RND \cite{prediction-error3}. To separate the effects of the boundary-estimation method and novelty function, we compare both methods using the same novelty measure. As shown in Fig.~\ref{novelty_compared_result}, ENTINEX outperforms NovelD in most environments, indicating that its improvement primarily arises from the entropic information-based approach.

\begin{figure*}[hbt!]
\centering
\begin{subfigure}[b]{\textwidth}
    \centering
    \includegraphics[width=0.7\textwidth]{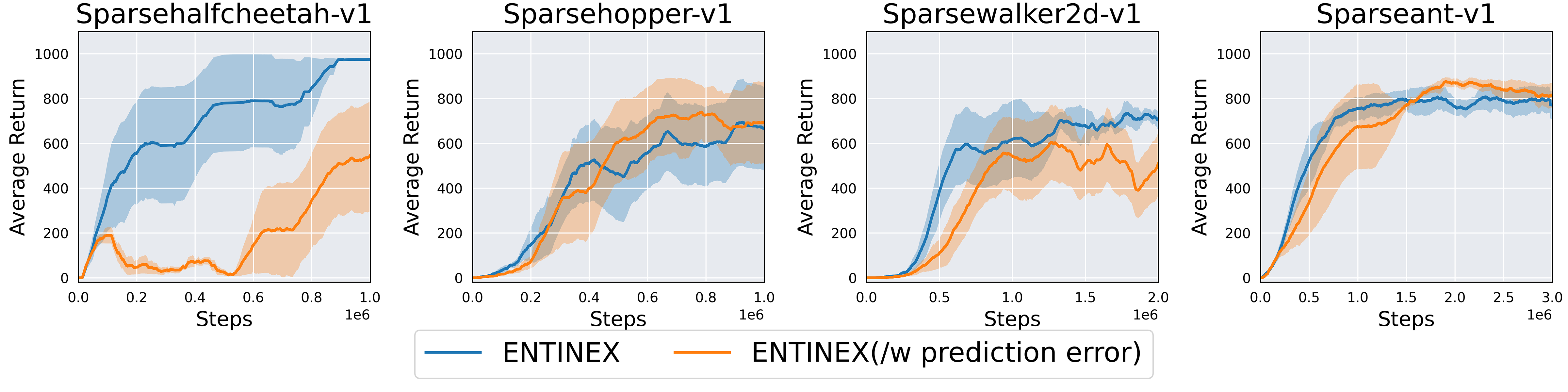}\label{novelty_func_eval_sparse}
\end{subfigure}
\hfill
\begin{subfigure}[b]{\textwidth}
    \centering
    \includegraphics[width=0.7\textwidth]{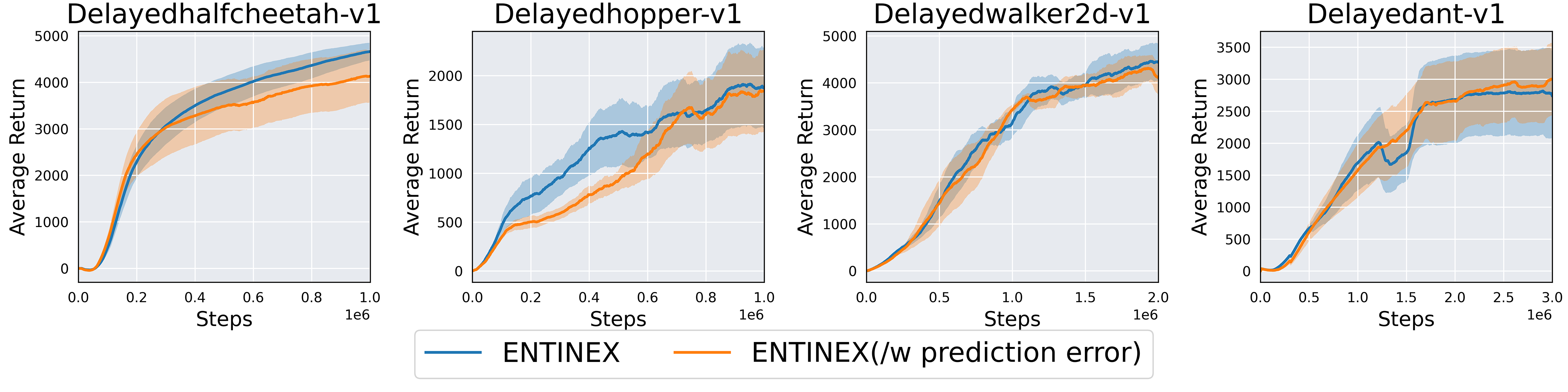}\label{novelty_func_eval_delayed}
\end{subfigure}
\caption{Performance evaluation of ENTINEX with two different novelty functions.}
\label{novelty_func_eval}
\end{figure*}

\subsubsection{Effect of the State Novelty Function}
ENTINEX can be combined with different state novelty measures. Although our default implementation uses the state-entropy estimator of RE3 \cite{re3}, we also evaluate ENTINEX with the prediction-error estimator of RND \cite{prediction-error3}. As shown in Fig.~\ref{novelty_func_eval}, both variants perform consistently across the evaluated environments, with the state-entropy variant achieving slightly better results. This suggests that ENTINEX is not restricted to a particular novelty function and may benefit from improved novelty estimators.

\subsubsection{Unsupervised Pre-training}
\begin{figure}[t]
    \begin{center}
    \vspace{-\baselineskip}
        \includegraphics[width=0.4\textwidth]{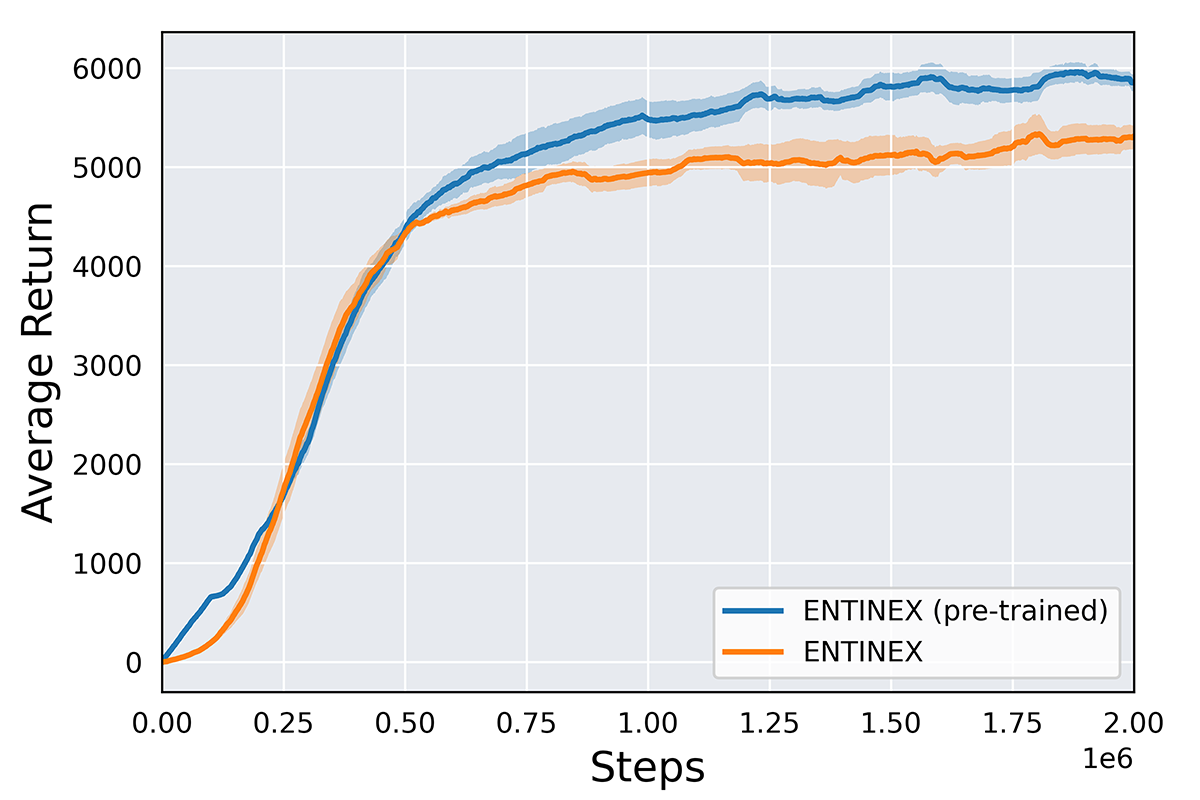}
    \end{center}
\caption{Performance evaluation to validate the effectiveness of unsupervised pre-training on the Walker2d environment.}
\label{unsupervised_eval}
\end{figure}

By rewarding states near the SND boundary, ENTINEX trains a policy to visit diverse regions of the state space. We evaluate whether this behavior improves downstream sample efficiency through unsupervised pre-training. Specifically, we pre-train the policy for 2M steps using only intrinsic rewards, save checkpoints throughout training, and fine-tune the selected pre-trained policy using the original task reward. As shown in Fig.~\ref{unsupervised_eval}, pre-training with ENTINEX improves sample efficiency and enables the agent to achieve higher average returns.

\section{Discussion}
We proposed ENTINEX, an entropic information-based exploration method that identifies the state-novelty distribution boundary and assigns intrinsic rewards to states near it. Experiments under sparse and delayed rewards show that ENTINEX consistently outperforms existing exploration methods, remains effective with different novelty functions, and improves the sample efficiency of off-policy RL through unsupervised pre-training.

ENTINEX may introduce training instability and additional computational overhead because it maintains multiple components, including a dynamics model and a state novelty estimator. Moreover, a rigorous theoretical characterization of the proposed approach remains for future work. Despite these limitations, the results demonstrate the potential of entropic boundary exploration and suggest that further improvements to its components may enhance performance.

\begin{credits}
\subsubsection{\ackname} This work was supported in part by the Institute of Information Communications Technology Planning Evaluation (IITP) funded by Korean Government under Grant 2022-0-00469, and in part by the BK21 FOUR(Connected AI Education \& Research Program for Industry and Society Innovation, KAIST EE, No. 4120200113769)

\end{credits}

\bibliographystyle{splncs04}
\bibliography{reference}

\end{document}